# Quasi-Static Analysis on Transoral Surgical Tendon-Driven Articulated Robot Units

Hojin Seo
George W. Woodruff School of
Mechanical Engineering
Georgia Institute of Technology
Atlanta, United States
hseo47@gatech.edu

Yeoun-Jae Kim
Biomedical Engineering Research
Center
Asan Institute for Life Sciences, Asan
Medical Center
Seoul, South Korea
lethkim1@gmail.com

Jaesoon Choi
Department of Biomedical Engineering
Asan Medical Center, University of
Ulsan College of Medicine
Seoul, South Korea
fides@amc.seoul.kr

Youngjin Moon*
Department of Convergence Medicine
University of Ulsan College of
Medicine
Seoul, South Korea
jacobian@amc.seoul.kr

*Abstract* — **Wire actuation in tendon-driven continuum robots enables the transmission of force from a distance, but it is understood that tension control problems can arise when a pulley is used to actuate two cables in a push-pull mode. This paper analyzes the relationship between angle of rotation, pressure, as well as variables of a single continuum unit in a quasi-static equilibrium. The primary objective of the quasi-static analysis was to output pressure and the analysis, given the tensions applied. Static equilibrium condition was established, and the bisection method was carried out for the angle of rotation. The function for the bisection method considered pressure-induced forces, friction forces, and weight. $\theta$ was 17.14°, and $p$ was 405.6 Pa when $T_l$ and $T_s$ were given the values of 1 N and 2 N, respectively. The results seemed to be consistent with the preliminary design specification, calling for further simulations and experiments.**

## I. Introduction

Transoral robotic surgery is a procedure to remove pharyngeal and laryngeal cancer with the use of robotic instruments and computer-enhanced systems to guide surgical tools [1]. Traditional transoral surgery is difficult to perform due to constrained workspace of upper airways and lack of dexterity of conventional instruments, but flexibility of wire-actuated continuum robotic devices is expected to improve distal dexterity in surgical environments [2]. Wire actuation enables the transmission of force from a distance, and the absence of motorized joints maximizes its flexibility [3]. These features have allowed several breakthroughs in recent years and wire-actuated continuum robots have been widely studied [4]. It is, however, understood that tension control problems can arise when a pulley is used to actuate two cables in a push-pull mode [5][6]. This paper analyzes the relationship between angle of rotation, pressure, as well as variables of a single continuum unit in a quasi-static equilibrium.

## II. Overview

The primary objective of the quasi-static analysis was to output $p$ and $\theta$, given $T_l$ and $T_s$, where $p$ is an average of the distribution of pressure applied by the first unit on the second unit, $\theta$ is angle of rotation on X-Z plane, $T_l$ and $T_s$ are the two tension forces applied to the robot units (See Fig. 1). The initial planar static equilibrium condition was established for the forces, as for the moment in the direction of the angle of rotation. The pressure-induced forces that describe the equilibrium conditions were expressed as integrands using $p$. Cable forces, frictions, and pressure-induced forces were used to rewrite the initial static equilibrium conditions. The bisection method was carried out using MATLAB after merging the three equations for static equilibrium to find the corresponding $\theta$ with the given inputs [7].

## III. Static equilibrium of a unit section

This paper investigates the static equilibrium of a unit with a fixed wire position assuming no elongation, no slack, and no friction between the wire and the unit section. Fig. 1 is a free body diagram on the vertical cross-section of a tilted unit connected to a different unit,

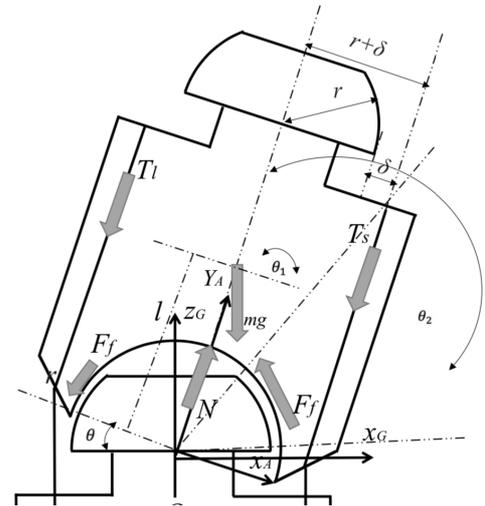

Fig. 1 Free Body Diagram of Unit Static Equilibrium

$$\Sigma F_{X_G}: +(-(T_l + T_s) + N)\sin\theta - F_{f_{X_G}} = 0 \quad (1)$$

$$\Sigma F_{Z_G}: +(-(T_l + T_s) + N)\cos\theta + F_{f_{Z_G}} = mg \quad (2)$$

$$\Sigma M_{Y_G}: (r+\delta)(T_s - T_l) - F_f r + mg\sin\theta\, l = 0 \quad (3)$$

where
- $F_f$  friction forces between the surface of the units;
- $l$  distance to the center of gravity;
- $r$  radius of the ball joint;
- $\delta$  distance between the edge of socket joint and the wire cable;
- $N$  pressure induced forces.

Tension forces ($T_l$ and $T_s$), friction forces ($F_f$), pressure-induced forces, and weight are acting on the tilted unit. With these forces, equations (1), (2), and (3) describe the static equilibrium of forces in the X direction, Z direction, and the equilibrium of the moment respectively.

## IV. BISECTION METHOD

The bisection method is a numerical method to find a root given a function and a range [8]. The function was created in terms of $\theta$ from the equilibrium equations (1) (2) and (3), after the equilibrium equations were rewritten with antiderivatives describing pressure-induced forces and friction. Pressure-induced forces expressed in antiderivatives of $p$ in the X direction and Z direction were expressed as $C_{F_{X_G}}$ and $C_{F_{Z_G}}$ respectively. The equivalent friction forces on the rotating surface were written as $C_{f_{X_G}}$ and $C_{f_{Z_G}}$. The antiderivatives were evaluated from the first rotational point, $\theta_1$, to the second rotational point, $\theta_2$, of a unit. After the equilibrium equations were rewritten, the average pressure distributed across the joint surface, $p$, was isolated from the expression of pressure-induced forces, tension forces and friction forces. $p$ was then expressed in quadratic formula, allowing the variables to be reduced to only $\theta$ by expressing the root of the quadratic in $\theta$. The equation was:

$$\frac{((T_s - T_l)(r + \delta) + mg\sin\theta l)}{\mu_s r \sqrt{C^2_{f_{X_G}} + C^2_{f_{Z_G}}}} = \frac{(T_l + T_s)\left(\sqrt{C_{F_{X_G}}^2 + C_{F_{Z_G}}^2} - mg\mu_s C_{f_{Z_G}}\right) - \sqrt{\left((T_l+T_s)\left(\sqrt{C_{F_{X_G}}^2+C_{F_{Z_G}}^2}-mg\mu_s C_{f_{Z_G}}\right)\right)^2 - \left(\left[\mu_s C_{f_{X_G}}^2 + \mu_s C_{f_{Z_G}}^2 - \left(C_{F_{X_G}}^2 + C_{F_{Z_G}}^2\right)\right]\right)\left((mg)^2 - (T_l+T_s)^2\right)}}{\left[\mu_s C_{f_{X_G}}^2 + \mu_s C_{f_{Z_G}}^2 - \left(C_{F_{X_G}}^2 + C_{F_{Z_G}}^2\right)\right]} \quad .(4)$$

$C_{F_{X_G}}$, $C_{F_{Z_G}}$, $C_{f_{X_G}}$, and $C_{f_{X_G}}$ were expressed in terms of $\theta$, as $l$, $r$, $\delta$, and mass of the unit, $m$, which were given by the preliminary design of robot module from a preceding paper [9]. From the preliminary design, $l$ is 0.008m, $r$ is 0.0078m, $\theta_1$ is 60°, and $\theta_2$ is 88.49°. $\mu_s$ is 1.05, as the continuum robot section was made out of aluminum [10]. Range of bisection is between 0° and 30°, as the robot could not physically bend more than 30°. Bisection method was carried out for the angle of rotation, $\theta$, when $T_l$ and $T_s$ were given the values of 1 N and 2 N, respectively.

## V. RESULTS

The bisection method was carried out using MATLAB, where it terminated when the root converged with an error of $1^{-10}$ for $\theta$. $\theta$ was 17.14°, and $p$ was 405.6 Pa. The derivatives of pressure-induced force in the X direction and Z direction, as well as all the friction forces expressed in terms of $p$ were checked for signs with $\theta$ and $p$. $\theta$ lied between 0 and 30°, as predicted from the preliminary design. Equation (4) is an exponentially growing equation as shown in Fig. 2.

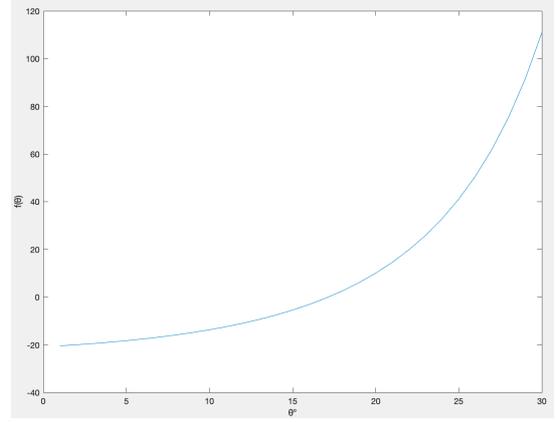

Fig. 2 Equation (4) for the bisection method plotted for $\theta$

## VI. CONCLUSIONS

This paper summarized a quasi-static analysis on a tendon-driven robot section by studying the conditions of static equilibrium and performing the bisection method on a function defined by $\theta$, angle of rotation. $\theta$ and $p$ seemed to be consistent with the preliminary design. The primary objective to output $p$ and $\theta$, given $T_l$ and $T_s$ appeared to be successful. $\theta$, however, seems to be finicky when $T_l$ and $T_s$ were varied. The results of this paper require further validation from simulations and experiments.